\documentclass{article}
\usepackage{times}

\usepackage{amssymb}
\usepackage{amsmath}
\usepackage{graphicx}
\usepackage{latexsym}
\usepackage{ marvosym }
\usepackage{algorithmic}
\usepackage[ ruled,vlined,linesnumbered]{algorithm2e}

\usepackage{multirow}

\def\closedx{\textsc{ClosedPattern}\xspace}
\def\coversize{\textsc{CoverSize}\xspace}

\newcommand{\items}{\mathcal{I}} 			%ensemble des items
\newcommand{\trans}{\mathcal{T}}				%jeu de données

\newcommand{\angx}[1]{{{\mbox{$\langle #1 \rangle$}}}}

\newcommand{\SDB}{\mathcal{D}}	

\newcommand{\I}{\mathcal{I}}

\newcommand{\lang}[1]{\mathcal{L}_\mathcal{#1}}

\usepackage{colortbl}
\usepackage{xcolor,colortbl}

\definecolor{Gray1}{gray}{0.95}
\definecolor{Gray2}{gray}{0.90}
\definecolor{Gray3}{gray}{0.85}
\definecolor{Gray4}{gray}{0.80}
\definecolor{Gray5}{gray}{0.35}
\definecolor{LightCyan}{rgb}{0.88,1,1}

\newcommand{\X}{P}	
\newcommand{\XX}{H}	
\newcommand{\Y}{T}	
\newcommand{\YY}{V}	

\newcommand{\itemset}{{\sc ItemSet}\xspace}	
\newcommand{\cp}{{\sc cp-ItemSet}\xspace}	
\newcommand{\lcm}{{\sc lcm}\xspace}	
\newcommand{\FCI}{{$\tt \#FCIs$}\xspace}	
\newcommand{\pplcm}{{\sc pp-lcm}\xspace}

\newcommand{\zoo}{{$\tt Zoo$}}
\newcommand{\primary}{{$\tt Primary$}}	
\newcommand{\vote}{{$\tt Vote$}}	
\newcommand{\chess}{{$\tt Chess$}}	
\newcommand{\mushroom}{{$\tt Mushroom$}}

\newcommand{\zooI}[2]{{$\tt Zoo\_#1\_#2$}}
\newcommand{\primaryI}[2]{{$\tt Primary\_#1\_#2$}}	
\newcommand{\voteI}[2]{{$\tt Vote\_#1\_#2$}}	
\newcommand{\chessI}[2]{{$\tt Chess\_#1\_#2$}}	
\newcommand{\mushroomI}[2]{{$\tt Mushroom\_#1\_#2$}}

\newtheorem{Example}{Example}

\SetKw{myproc}{procedure}
\SetKw{myfunc}{function}
\SetKw{Inp}{in}
\SetKw{Outp}{out}
\SetKw{InOutp}{in out}
\SetKw{F}{false}
\SetKw{lAnd}{and}

\usepackage{etoolbox}
\newbool{seeAll}
\usepackage[normalem]{ulem}
\usepackage{multicol}
\def\trackingLevel{2}

\newcommand{\cb}[1]{\comment{blue}{CB}{#1}}
\newcommand{\nad}[1]{\comment{red}{NL}{#1}}
\newcommand{\mm}[1]{\comment{red}{MM}{#1}}

\newcommand{\comment}[3]{\ifnumcomp{\trackingLevel}{=}{2}{{\color{#1}[\bf\footnotesize{{#2: \textit{#3}}}]}}{}}

\begin{document} 
%\title{Constraint-Based Itemset Mining Using Constraint Programming}
\title{User's Constraints in Itemset Mining}
\author{Christian Bessiere\\ CNRS, University of Montpellier\\ France\\bessiere@lirmm.fr \and 
Nadjib Lazaar\\University of Montpellier\\ France\\lazaar@lirmm.fr  \and 
~~~~Yahia Lebbah\\ ~~~~~LITIO,  University of Oran 1\\~~~~~Algeria\\~~~~~lebbah.yahia@univ-oran.dz  \and
~~~~~Mehdi Maamar\\ ~~~~~CRIL, University of Artois\\~~~~~France\\ ~~~~~maamar@cril.fr 
}
\date{}
\maketitle

\begin{abstract}
Discovering significant itemsets is one of the fundamental tasks in data mining.
It has recently been shown that constraint programming  is a flexible 
way to tackle  data mining tasks.
With a constraint programming approach, 
we can easily express and efficiently answer queries with user's constraints on itemsets.
However, in many practical cases queries also 
involve user's constraints on the dataset itself. 
For instance, in a dataset of purchases, 
the user may want to know which itemset is frequent 
and the day at which it is frequent. 
This paper presents a general constraint programming model able to 
handle any kind of query on the dataset for itemset mining.
%A sentence on the implementation and the XP. Conclusion.

\end{abstract}

%%%%%%%%%%%%%%%%%%%%%%%%%%%%%%%%%%%%%%%%%%%%%%%%%%%%%%%%%%%%%%%%%%%%%%%%%%%%%%%%%%%%
%%%%%%%%%%%%%%%%%%%%%%%%%%%%%%%%%%%%%%%%%%%%%%%%%%%%%%%%%%%%%%%%%%%%%%%%%%%%%%%%%%%%
%%%%%%%%%%%%%%%%%%%%%%%%%%%%%%%%%%%%%%%%%%%%%%%%%%%%%%%%%%%%%%%%%%%%%%%%%%%%%%%%%%%%

\section{Introduction}
\label{sec:intro}

People have always been  interested in analyzing phenomena from data 
by looking for significant itemsets.
This task became easier and accessible for big datasets thanks to 
computers, and thanks to the development of specialized algorithms 
for finding frequent/closed/... itemsets.
Nevertheless, looking for itemsets with additional user's constraints 
remains a bottleneck  nowadays.
According to  \cite{Wojciechowski}, there are three ways to handle 
user's constraints in an itemset mining  problem. 
We can use a pre-processing step that   restricts the dataset to only transactions 
that satisfy the constraints. 
%	\nad{v55:  We have to update this part, I forgot to talk also on the fact that we can have a pre-processing to remove useless columns (items)}. 
Such a technique quickly becomes infeasible when there is a large number of sub-datasets satisfying the user's constraints.
We can integrate the filtering of the user's constraints into the 
specialized data mining process in order to extract  only the itemsets satisfying 
the constraints. Such a technique requires the development of a new algorithm 
for each new itemset mining  problem with user's constraints.  
We can sometimes use a post-processing step to filter out the itemsets violating 
the user's constraints. Such a brute-force technique does not apply to all kinds of 
constraints and is computationally 
infeasible when the  problem  without the user's constraints has many solutions.

In a recent line of work \cite{de2008constraint,CI-KHIARI-10b,Lazaar2016,DBLP:journals/constraints/KemmarLLBC17,coverSize17},
constraint programming (CP) has been used as 
a declarative way to solve data mining problems. 
Such an approach has not competed yet with state of the art 
data mining algorithms~\cite{DBLP:conf/sdm/ZakiH02,DBLP:conf/dis/UnoAUA04} 
%in terms of CPU time 
for simple queries. 
Nevertheless,  the advantage of the CP 
approach is to be able to add extra (user's) constraints in the model 
so as to generate only \emph{interesting} itemsets at no other implementation cost. 
%In a pure data mining approach, the algorithm is just able to generate 
%all frequent/closed/... itemsets 
%and post-filter the  ones violating the user's constraints, 
%which can be far too expensive. 

The weakness of the CP approach is that 
the kind of  user's constraints that can be 
expressed has not been  clarified. 
%
%The CP approach works well if the user wants to 
%specify that she is only interested in itemsets that contain a given item $i$ or itemsets 
%of a given minimal length. 
%The solver will return the FCIs  containing $i$ or . 
%
With  the global 
constraint for frequent closed itemsets (FCIs) proposed  
in \cite{Lazaar2016}, if the  user 
%
%can easily specify that 
%she is only interested in itemsets that contain a given item $i$, 
%or in itemsets of a given minimal size $k$.  
%The solver will return FCIs containing $i$ or 
%itemsets with at least $k$ items. 
%However, not all kinds 
%of constraints on the items are compatible with the closedness property involved in FCIs. 
%If the user 
is only interested in itemsets 
{not} containing item $i$, 
%or in itemsets of size {smaller} than a given $k$, 
this constraint will interfere  with the closedness property 
so that some itemsets are lost.   
(See \cite{BonchiL04} for a characterization of this issue). 
In \cite{coverSize17}, the issue is addressed by relaxing the 
closedness property, but this can lead 
to a dramatic increase of the number of solution itemsets. 
Another  important issue is  that the user may be  
interested in mining only in  \emph{transactions} containing a given item $i$, or 
in transactions corresponding 
to customers having spent less than 100\EUR\ in her shop. 
%In the current semantics, the CP solver will return any 
%closed frequent itemset not containing the item $e$ or costing less than 100\EUR\ 
%even if this itemset is seldom included in transactions without the item $e$ or in 
%transactions of less than 100\EUR. 
None of the current CP models is able to catch such kind of constraints. 
As  specialized approaches, we need to follow a generate-and-test process with an ad-hoc algorithm
able to generate datasets with transactions including the item $e$ or transaction  
costing less than 100\EUR.

%This can  somewhat be embarrassing to return rare itemsets in a frequent 
%pattern mining task. 
%What the user could have expected is to be returned patterns that are frequent 
%in the subset of transactions not containing $E$ or costing less than 100\EUR.

The contribution of this paper is two-fold. 
%\begin{enumerate}
%	\item 
We present a first  classification of the user's constraints w.r.t. 
	 which and where the itemsets are extracted. 
%	\item 
and we propose a generic CP model in which we can capture any kind of user's constraints.
%\end{enumerate}
It  is the first  system  able to mine 
closed itemsets in the presence of any kind of user's constraints. 
(In \cite{BonchiL04}, the algorithm tackles monotone and anti-monotone constraints only.)

The paper is organized as follows. Section \ref{sec:background} presents the background. 
% in data mining and constraint programming. 
In Section \ref{sec:taxonomy} we present a taxonomy of the types of user's constraints. 
% that can be useful in itemsets mining. 
In Section \ref{sec:model}, we present a CP model able to capture all these user's constraints. 
Section~\ref{sec:cases} gives some case studies that can be expressed using our CP model.
%, in section 5 we execute our proposed filtering rules on a concrete example, experiment results are represented in section 6, 
%We conclude in Section 6. %\mm{update when sections are stable.}
Section \ref{sec:expes} reports experiments. 
%Finally, we conclude and draw some perspectives.

\section{Background}
\label{sec:background}

\subsection{Itemset mining}
Let $\mathcal{I}=\{1, \ldots, n\}$ be a set of $n$ \textit{item} indices
and $\mathcal{T} = \{1,\ldots,m\}$ a set of $m$ \textit{transaction} indices. 
An itemset $P$ is a  subset of $\mathcal{I}$. 
The set of itemsets is $\lang{I}= 2^{\mathcal{I}} \backslash \emptyset$. 
A transactional dataset is a set  $\mathcal{D} \subseteq \mathcal{I} \times \mathcal{T}$. 
%Let 
%$D_{i*}$ be the transaction of index $i$. 
% (horizontal representation) and
%$D_{*j}$ be the v-item\cb{what is that?} of index $j$ (vertical representation).  
%$D_{ij}=1$ if $j\in D_{i*}$, $0$ otherwise. 
%\nad{to check if these two notations are really used in the paper}
%{In our context, we can ask for itemsets extracted 
%from a particular part of the datatset. 
A sub-dataset is a subset of $\mathcal{D}$ obtained by removing columns (items) 
and/or rows (transactions). 
The set of possible sub-datasets is denoted by $\lang{D}$. 
%element of $(2^{\mathcal{T}}\backslash \emptyset)\times (2^{\mathcal{I}}\backslash \emptyset)$.
%
The cover $cover(D,P)$ of an itemset $P$ in a  sub-dataset $D$ 
%denoted by $cover(D,P)$, 
is the set of transactions in $D$ containing $P$. 
The frequency 
%$frequency(D,P)$ 
of an itemset $P$ in $D$,   
is the  ratio $\frac{|cover(D,P)|}{|cover(D,\emptyset)|}$.

%Itemset mining aims at extracting all itemsets $X$ of $\mathcal{L}_{\mathcal{I}}$ 
%satisfying a query $Q(X)$ (i.e., conjunction of constraints). 
%The common examples are the frequency measure leading to the minimal frequency constraint, the closed itemset constraint, the maximal itemset constraint, itemset size constraint, etc.

\subsection{Constraint programming (CP)}
A constraint program is defined by a 
set of variables $X = \{X_1,$ $ \ldots,$ $ X_n\}$, 
where $D_i$ is
the set of values that can be assigned to $X_i$, 
and a
finite set of constraints $\mathcal{C}$. 
Each constraint $C(Y) \in \mathcal{C}$ expresses a relation over a 
subset $Y$ of variables $X$.
The task  is to find  assignments ($X_i = d_i$) with $d_i \in D_i$
for $i = 1, \ldots, n$, such that all constraints are satisfied. 

%\adding{example!!}

\subsection{CP models for  itemset mining}\label{CPIM}
%Different CP models have been proposed for data mining tasks. 
%In this section we discuss the different CP approaches for  
%itemset mining problem. 

In \cite{de2008constraint,guns2011itemset}, De Raedt {\it et al.} have proposed a first CP model for itemset mining. 
They showed how some constraints (e.g., frequency and closedness) 
can be formulated using CP. 
This model uses two sets of Boolean variables: 
(1) item variables $\{X_1, X_2,..., X_n\}$
where $(X_i = 1)$ iff the extracted itemset $P$ contains $i$; 
(2) transaction variables 
$\{T_1, T_2,..., T_m\}$ where $(T_t =1)$ iff $(P \subseteq t)$.
The relationship between $P$ and $T$ is modeled by $m$ reified
$n$-ary constraints. 
The minimal frequency constraint 
%is encoded by $n$ reified $m$-ary constraints.
and the closedness constraint are also  encoded by $n$-ary and $m$-ary reified constraints.

Recently, global constraints have been proposed to model and solve efficiently data mining problems. 
The \closedx global constraint  in
\cite{Lazaar2016}  compactly encodes both the minimal 
frequency and the closedness constraints. 
This global constraint does not use reified constraints. 
It is defined only on item variables. 
The  filtering algorithm ensures domain consistency in a polynomial 
time and space complexity. 
The \coversize{} global constraint  in \cite{coverSize17} 
uses a reversible sparse bitset data structure to compute 
the subset of transactions that cover an itemset. The  filtering 
algorithm computes a lower and an upper-bound on the frequency.  

%The global constraint \closedx$_{\mathcal{D},\theta}(P)$ holds iff  
%there exists an assignment $\sigma=\angx{d_1, ..., d_{n}}$ of
%variables $P$ s.t. $freq_{\mathcal{D}}(\sigma)\geq \theta$ and $closed_{freq}(\sigma)$.  

%%%%%%%%%%%%%%%%%%%%%%%%%%%%%%%%%%%%%%%%%%%%%%%%%%%%%%%%%%%%%%%%%%%%%%%%%%%%%%%%%%%%
%%%%%%%%%%%%%%%%%%%%%%%%%%%%%%%%%%%%%%%%%%%%%%%%%%%%%%%%%%%%%%%%%%%%%%%%%%%%%%%%%%%%
%%%%%%%%%%%%%%%%%%%%%%%%%%%%%%%%%%%%%%%%%%%%%%%%%%%%%%%%%%%%%%%%%%%%%%%%%%%%%%%%%%%%

\section{User's Constraints Taxonomy}
\label{sec:taxonomy}

For an itemset mining task we aim at extracting all itemsets $P$ of $\mathcal{L}_\mathcal{I}$ 
satisfying a query $Q(P)$ that is a conjunction of (user's) constraints. 
The set $Th(Q)=\{ P\in \mathcal{L}_\mathcal{I}\mid Q(P) \}$ 
is called \textit{a theory} \cite{MannilaT97}. 
Common examples of user's constraints on extracted itemsets 
are frequency, closure, maximality, etc. 
%The semantics of such constraints is caught by a query predicate $Q(P)$ expressed on itemsets.
However, it may be desirable for a user to 
ask for itemsets extracted from  particular parts 
%$\lang{D}$ 
of the dataset. 
%If all such sub-datasets 
%can be located using a preprocessing step, 
%the mining process can act in two steps: isolate the sub-datasets; extract itemsets.
%But, isolating all sub-datasets that satisfy the user's requirements can  be  challenging.
In the  general case, 
a query predicate, denoted by $Q(D,P)$, is expressed both 
on the itemsets $P$ it returns and on the sub-datasets {$D\in \lang{D}$} 
on which it  mines.
The extracted elements forming a theory are now pairs: 
\begin{center}
	$Th(Q)=\{ (D,P)\mid D\in \lang{D} \wedge P\in \lang{I} \wedge Q(D,P)\}$. 
\end{center}

To make the description of our user's constraints 
taxonomy less abstract, we suppose  a categorization 
of items and transactions.
Items are products  belonging  to $k$ categories (e.g., 
food, electronics, cleaning, etc), denoted by 
$\mathcal{I}=$ $\{\mathcal{I}_1,$ $\ldots,$ $\mathcal{I}_k\}$.
Transactions are categorized into $v$ 
categories of customers (e.g., categories based on age/gender criteria), denoted by 
$\mathcal{T}=\{\mathcal{T}_1,\ldots,\mathcal{T}_v\}$.
Example \ref{ex:table} presents the 
running example (with categories) that will be used 
to illustrate 
each of the types of user's constraints we present 
in this section.

\begin{Example}\label{ex:table}
The dataset $\SDB_1$ involving 9 items and 6 transactions 
is displayed in Table \ref{table:example}. 
Items  belong to three categories: $\{A,B\}$, $\{C,D,E\}$ and $\{F,G,H,K\}$. 
Transactions belong to three categories as well: $\{t_1,t_2\}$, $\{t_3,t_4\}$ and $\{t_5,t_6\}$.

	%%%%%%%%%%%%%%%%%%%%%%%%%%%%%%%%%%%%%%%%%%%%%%%%%%%%%%%%%%%%%%%%%%%%%%%%%%%%
	\begin{table}[tb]
		\caption{ A transaction dataset $\SDB_1$. }
		\label{table:example}
			\centering
				\scalebox{0.95}{
					\begin{tabular}{cc|cc|ccc|cccc}
						\hline
						 & \multicolumn{8}{c}{Items} \\ 
						\multicolumn{2}{c|}{Trans.} & \multicolumn{2}{c|}{$\mathcal{I}_1$} & \multicolumn{3}{c|}{$\mathcal{I}_2$}& \multicolumn{3}{c}{$\mathcal{I}_3$}\\ 
						\hline
	\rowcolor{Gray3}									& $t_1$ &   & B & C &   &   &   & G & H & K\\ 
	\rowcolor{Gray3} \multirow{-2}{*}{$\mathcal{T}_1$}		& $t_2$ & A &   &   & D &   &   & G &  & K \\ 
	\rowcolor{LightCyan}								&$t_3$ & A &   & C & D &   &   &   & H & \\ 
	\rowcolor{LightCyan} \multirow{-2}{*}{$\mathcal{T}_2$}	&$t_4$ & A &   &   &   & E & F &   &  & \\ 
	\rowcolor{Gray3}									&$t_5$ &   & B &   &   & E & F &   &  & \\ 
	\rowcolor{Gray3} \multirow{-2}{*}{$\mathcal{T}_3$}		&$t_6$ &   & B &   &   & E & F & G &  & K \\ 
						\hline
					\end{tabular}
				}

	\end{table}
	%%%%%%%%%%%%%%%%%%%%%%%%%%%%%%%%%%%%%%%%%%%%%%%%%%%%%%%%%%%%%%%%%%%%%%%%%%%%%%
	
\end{Example}

	\begin{figure}[tb]
		\centering 
		
		\includegraphics[scale=0.3]{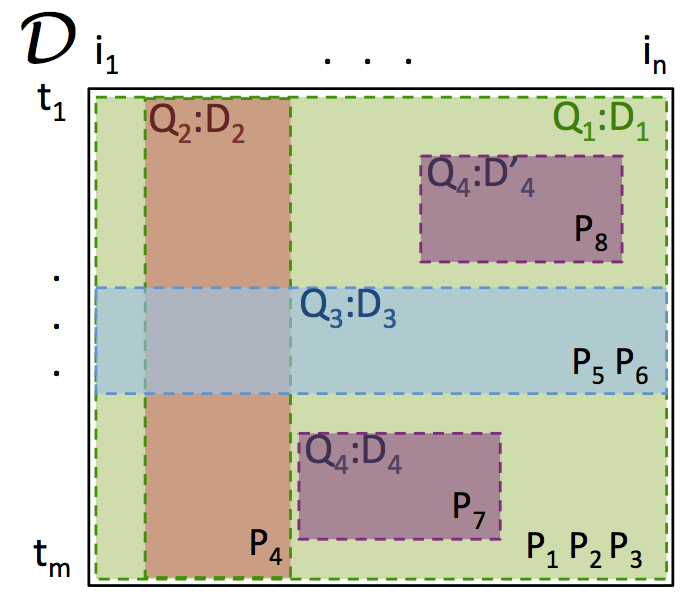}
		\caption{Queries on dataset $\mathcal{D}$.
		}		\label{fig:example}
	\end{figure}

\subsection{User's constraints on itemsets}\label{sec:q1}
		
When the user comes with constraints only on the nature of the itemsets to extract, the query, $Q_1$, is equivalent to a standard itemset mining task. 
%No constraint occurs on the dataset $\SDB$, that is, 
We mine on the whole  dataset. 
Figure \ref{fig:example} graphically illustrates this. 
The  itemsets that are  solution for $Q_1$ 
(i.e., $P_1$, $P_2$ and $P_3$) are extracted from  $D_1=\SDB$.

An example of such a query where user's constraints are expressed only on itemsets 
is the query  $Q_1$ asking for  FCIs:
$$Q_1(D,P)\equiv freqent(D,P,\theta) \wedge closed(D,P)$$
%\cb{v37: why D appears?}  
where $frequent(D,P,\theta)$ and $closed(D,P)$ 
%\cb{take care of notations. See background once stable} 
are predicates 
expressing user's constraints on the frequency (with a minimum frequency  $\theta$) 
and the closure of an itemset $P$ in~$D$, where $D$ is  $\SDB$ in this case. 
The query $Q_1$ on the dataset $\SDB_1$ of Table \ref{table:example} 
with a minimum frequency $\theta\geq50\%$ returns $A$, $B$, $EF$ and $GK$ as FCIs.

As a second example of such a query on itemsets 
%let us take a dataset $\mathcal{D}$ with $k$  item categories.
the user can ask a query  $Q'_{1}$ 
where the extracted itemsets 
are FCIs and the items are taken from at least 
$lb$  and at most $ub$ categories:  % with $lb\leq ub\leq k$:
$$Q'_1(D,P)\equiv Q_1(D,P) \wedge atLeast(P,lb) \wedge atMost(P,ub)$$
where 
$atLeast(P,lb)$ and $atMost(P,ub)$  are user's constraints ensuring 
that the itemset $P$ overlaps between $lb$ and $ub$ categories of items. 
The query $Q'_1$ on the  dataset $\SDB_1$ of Table \ref{table:example} 
with  $lb=ub=2$ and  minimum frequency $\theta=50\%$ 
only returns $EF$. It does not return $A$, $B$, and $GK$ 
because each of these itemsets belongs to a single category. 
%According to what the user is asking for, note that for instance the itemset $GK$ is an FCIs on $D_1$ (i.e., it satisfies $Q_1$), but it does not satisfy $Q'_1$ where the two items are in the same category (i.e., the {\em atLeast} and the {\em atMost} predicates are violated). 

\subsection{User's constraints on items}\label{sec:q2}
In addition to  constraints on itemsets, the user may want to put constraints on the items themselves. 
Such constraints are constraints on the dataset. 
They specify  on which items/columns the mining will occur. 
In Figure \ref{fig:example}, constraints on items 
lead the query, $Q_2$, to mine on the sub-dataset 
$D_2$ satisfying constraints on items.  
%from which  we extract the itemset $P_4$ satisfying 
%the  constraints on itemsets. 

As an example, %let us take a dataset $\SDB$ with $k$ item categories.
the user can ask a query 
{$Q_2$},
where the extracted itemsets are FCIs of
sub-datasets containing at least $lb_I$ categories of items  
and at most $ub_I$ categories:  
%with $lb_I\leq ub_I\leq k$: 
$$Q_2(D,P)\equiv Q_1(D,P) \wedge atLeastI(D,lb_I) \wedge atMostI(D,ub_I)$$
where $atLeastI(D,lb_I)$ and $atMostI(D,ub_I)$ are user's constraints ensuring 
that the dataset  $D$ contains between $lb_I$ and $ub_I$ categories of items. 
As opposed to $Q_1$ and $Q'_1$, 
{$Q_2$} 
seeks itemsets in sub-datasets satisfying a property on their items.
The query $Q_2$  on the  dataset $\SDB_1$ of Table \ref{table:example} 
with  $lb_I=ub_I=2$ and  minimum frequency $\theta=50\%$ 
returns 
%: 
%\begin{itemize}
%			\item   
$A$, $B$ and $E$ on $\mathcal{I}_1+\mathcal{I}_2$,
%			\item   
$A$, $B$, $F$ and $GK$ on $\mathcal{I}_1+\mathcal{I}_3$,
%			\item  
$EF$ and $GK$ on   $\mathcal{I}_2+\mathcal{I}_3$. 
%		\end{itemize}	

\subsection{User's constraints on transactions}\label{sec:q3}

The user may also want to put constraints on transactions. Such constraints  
determine on which transactions/rows the mining will occur.  
%Constraints on items were specifying on which items/columns to mine. 
%Constraints on transactions specify on which transactions/rows to mine. 
In Figure \ref{fig:example}, 
constraints on transactions lead the query, 
$Q_3$, to mine on the subset  $D_3$ of transactions. 
%from which we extract the itemsets  $P_5$ and $P_6$.   

As an example, 
%let us take a dataset $\SDB$ with transaction belonging to  $k$  categories of customers.
the user can ask a query 
{$Q_3$},
where the extracted itemsets 
are FCIs on at least $lb_T$  and at most $ub_T$ 
categories: 
% with $lb_T\leq ub_T \leq v$:
$$Q_3(D,P)\equiv Q_1(D,P) \wedge atLeastT(D,lb_T) \wedge atMostT(D,ub_T)$$
where
$atLeastT(D,lb_T)$ and $atMostT(D,ub_T)$ are user's constraints ensuring 
that the dataset  $D$ contains between $lb_T$ and $ub_T$ categories of transactions. 
%As opposed to $Q_1$, $Q'_1$, and $Q_2$, 
%{$Q_3$}
%seeks itemsets in sub-datasets satisfying a property on their transactions.
The query $Q_3$ on the  dataset $\SDB_1$ of Table \ref{table:example} 
with   $lb_T=ub_T=2$ and  minimum frequency $\theta=50\%$ 
returns:
%	\begin{itemize}		
%			\item  
$A$, $AD$, $CH$ and $GK$ on $\mathcal{T}_1+\mathcal{T}_2$,
%			\item 
$B$, $BEF$, $BGK$ and $GK$ on $\mathcal{T}_1+\mathcal{T}_3$, 
%			\item  
$A$, $BEF$ and $EF$ on $\mathcal{T}_2+\mathcal{T}_3$. 
%		\end{itemize}	

\subsection{User's constraints on items and transactions}\label{sec:q4}

Finally, the user may want to put constraints on both items and transactions. 
In Figure \ref{fig:example}, such constraints 
lead the query, $Q_4$, 
to mine on 
%, to return  particular itemsets  $P_7$ and $P_8$,   extracted from two different parts 
$D_4$ and $D'_4$. 
% of the dataset satisfying constraints on both items and transactions. 

%Consider our example where items and transactions  respectively belong to $k$ categories of products and $v$ categories of customers.
The user can ask a query 
{$Q_4$}, 
where the extracted itemsets are FCIs 
on at least $lb_I$  and at most $ub_I$ categories of items 
and on at least $lb_T$  and at most $ub_T$ categories of transactions:
$$Q_4(D,P)\equiv Q_2(D,P) \wedge Q_3(D,P)$$
The query $Q_4$ on the  dataset $\SDB_1$ of Table \ref{table:example} 
with  $lb_I=ub_I=lb_T=ub_T=2$ and  minimum frequency $\theta=50\%$ 
will have to explore nine possible sub-datasets in which to look 
for frequent closed itemsets:
\begin{table}[h!]
	\centering
	\def\arraystretch{1.5}
	\scalebox{0.85}{
	\begin{tabular}{c|c|c|c|}
		
	\multicolumn{1}{c|}{}	&$\mathcal{I}_1+\mathcal{I}_2$ & $\mathcal{I}_1+\mathcal{I}_3$ & $\mathcal{I}_2+\mathcal{I}_3$ \\ \hline
$\mathcal{T}_1+\mathcal{T}_2$	&	$A$, $AD$, $C$ & $A$, $GK$, $H$	& $CH$, $D$, $GK$\\ 
$\mathcal{T}_1+\mathcal{T}_3$	& $B$, $BE$	& $B$, $BF$, $GK$, $BGK$	& $EF$, $GK$\\ 
$\mathcal{T}_2+\mathcal{T}_3$	& $A$, $BE$, $E$ & $A$, $BF$, $F$	&	$EF$\\ \hline
	\end{tabular}
}
\end{table}

%
%\subsection{Complexity}
%In classic itemset mining mining,  where queries are of type 
%$Q(P)$ with only user's constraints on itemsets, the search 
%space contains $2^n$ candidate itemsets (e.g., $Q_1$ and 
%{$Q'_1$}). 
%If the user provides constraints on items, there are $2^{n}$ possible 
%sub-datasets defined by subsets of items, on which itemsets are extracted, 
%which gives a search space of  $2^{n}\times 2^n$ (e.g., {$Q_2$}). 
%If the user provides constraints on transactions, 
%the size of the search space becomes  $2^{m}\times 2^n$ (e.g., {$Q_3$}).
%Once the user expresses constraints both on items and transactions, 
%the size of the search space becomes $2^{(n+m)}\times 2^n$ (e.g., {$Q_4$}).

\subsection{A simple illustration: Where Ferrari cars are frequently bought?}
\label{sec:ferrari}
Consider a  dataset of cars purchases in France, where each transaction/purchase 
contains items representing  the city, the department, and the region where the purchase 
was performed. (City/department/region is the way France is administratively 
organized.)
The user may be interested in finding where 
(city, department or region) more than 10\% of the purchases are Ferrari cars. 
This can be done by the query:
\begin{align*}
	RQ(D,P)\equiv &freqent(D,P,10\%) \wedge (Ferrari\in P)\wedge \\
	&(Reg(D) \vee Dep(D)\vee City(D))
\end{align*}
where
$Reg(D)$,  $Dep(D)$ and $City(D)$ are user's constraints ensuring 
that the dataset  $D$ corresponds to one of the administrative entities of France.

%%%%%%%%%%%%%%%%%%%%%%%%%%%%%%%%%%%%%%%%%%%
%%%%%%%%%%%%%%%%%%%%%%%%%%%%%%%%%%%%%%%%%%%
% NL: we will see after if we have enough space to add RQ2
%%%%%%%%%%%%%%%%%%%%%%%%%%%%%%%%%%%%%%%%%%%
%%%%%%%%%%%%%%%%%%%%%%%%%%%%%%%%%%%%%%%%%%%
%%%%%%
%%%%%%\subsection{A simple illustration 2 : Area and Surface of a dataset} 
%%%%%%\mm{i don't know if we really need this example !}
%%%%%%\cb{v37: not edited}
%%%%%%
%%%%%%\textcolor{blue}{
%%%%%%Let us take a dataset $\mathcal{D}$ where items  are categorised (e.g., food, electronics, cleaning...).
%%%%%%That is, we have $\mathcal{I}$ partitioned into $k$ categories $\mathcal{I}_1,\ldots \mathcal{I}_k$.
%%%%%%The dataset is  fed at each time a customer makes a transaction, which leads us to a time ordered dataset.
%%%%%%A user wants to know the category of items that customers most often purchase and the corresponding time slot of the day.
%%%%%%Here, the user can express a threshold $\alpha$ on the quantity of purchased items and a minimum/maximum size of the returned time slot:}
%%%%%%\cb{this example is not described precisely enough}
%%%%%%
%%%%%%\textcolor{blue}{
%%%%%%\begin{align*}
%%%%%%&RQ_2(D,P)\equiv Density(D,\rho ) \wedge Area(D,\alpha)
%%%%%%\end{align*}
%%%%%%}

\section{A General CP Model for Itemset Mining}
\label{sec:model}

%In this section, w
We present \textsc{ItemSet},
a CP model for itemset mining taking into account any 
type of user's constraints presented in Section \ref{sec:taxonomy}.

\subsection{Variables}
$\X$, $\Y$, $\XX$  and $\YY$  are Boolean vectors to encode:

\begin{itemize}
	\item $\X=\angx{\X_1,\ldots,\X_n}$: the  itemset we are looking for. For each item $i$, 
	the Boolean variable $\X_i$ represents whether $i$ is in the extracted itemset.  
	
	\item $\Y=\angx{\Y_1,\ldots,\Y_m}$: the transactions that are covered by the extracted itemset.
	
	\item $\XX=\angx{\XX_1,\ldots,\XX_n}$: The items in the sub-dataset where the mining will occur. $\XX_i=0$ means that the item/column $i$ is ignored.
	
	\item $\YY=\angx{\YY_1,\ldots,\YY_m}$: The transactions in the sub-dataset where the mining will occur. $\YY_j=0$ means that the transaction/row  $j$ is  ignored.
	 
\end{itemize}

$\angx{\XX,\YY}$ circumscribes  the sub-dataset used to extract the itemset. 
The CP solver searches in different sub-datasets, backtracking from a  sub-dataset and  branching on another.
$\angx{\X,\Y}$  represents the  itemset we are looking for, and its coverage in terms of transactions.

\subsection{Constraints} 
%The vocabulary of variables $\angx{\X,\Y,\XX,\YY}$ allows the (user's) constraints to be easily and concisely expressed.
Our generic CP model consists of three sets of constraints:
\[
\textsc{ItemSet}(\X,\XX,\Y,\YY)=\begin{cases}
\textsc{DataSet}(\XX,\YY)\\
\textsc{Channeling}(\X,\XX,\Y,\YY)\\
\textsc{Mining}(\X,\XX,\Y,\YY)
\end{cases}
\]
%\begin{itemize}
%	\item 
$\textsc{DataSet}(\XX,\YY)$ is the set of constraints 	that  express 
user's constraints on items (i.e., $\XX$) and/or transactions (i.e., $\YY$). 
This  set of constraints circumscribes the sub-datasets. 
%	
%	\item 

$\textsc{Channeling}(\X,\XX,\Y,\YY)$ is the set of channeling constraints 
that express the relationship between the two sets of variables $\angx{\X,\Y}$ 
and $\angx{\XX,\YY}$:
\begin{align*}
\XX_i = 0 \Rightarrow \X_i = 0 	\\
\YY_j = 0 \Rightarrow \Y_j = 0 
\end{align*}
These constraints guarantee that if an item (resp. a transaction) is not 
part of  the mining process, it will not be part of the extracted itemset (resp. the cover set). 
%
%	
%	\item 

$\textsc{Mining}(\X,\XX,\Y,\YY)$ is the set of constraints that express the (user's) constraints on itemsets such as frequency, closedness, size, and more sophisticated user's constraints.
%\end{itemize}

%%%%%%%%%%%%%%%%%%%%%%%%%%%%%%%%%%%%%%%%%%%%%%%%%%%%%%%%%%%%%%%%%%%%%%%%%%%%%%%%%%%%
%%%%%%%%%%%%%%%%%%%%%%%%%%%%%%%%%%%%%%%%%%%%%%%%%%%%%%%%%%%%%%%%%%%%%%%%%%%%%%%%%%%%
%%%%
%%%%%%%%%%%%%%%%%%%%%%%%%%%%%%%%%%%%%%%%%%%%%%%%%%%%%%%%%%%%%%%%%%%%%%%%%%%%%%%%%%%%
%%%%%%%%%%%%%%%%%%%%%%%%%%%%%%%%%%%%%%%%%%%%%%%%%%%%%%%%%%%%%%%%%%%%%%%%%%%%%%%%%%%%
%%%%%%%%%%%%%%%%%%%%%%%%%%%%%%%%%%%%%%%%%%%%%%%%%%%%%%%%%%%%%%%%%%%%%%%%%%%%%%%%%%%%

\section{\textsc{ItemSet} Model: Cases Studies}
\label{sec:cases}

In this section, we illustrate our CP model  \textsc{ItemSet} on the queries detailed in Section \ref{sec:taxonomy}.
%We must bear in mind that f
For each query, user's constraints can be written in 
the \textsc{DataSet} and/or \textsc{Mining} parts of the \textsc{ItemSet} model. 
\textsc{Channeling} remains  unchanged.

\subsection*{Query $Q_1$}
For query $Q_1$, we have user's constraints only on itemsets.
That is, the mining process will occur on the whole set of transactions.
For such a case, we have:
\[
\textsc{DataSet}(\XX,\YY)=\begin{cases}
\forall i \in \mathcal{I} : \XX_{i} = 1\\
\forall j \in \mathcal{T} :\YY_{j} = 1
\end{cases}
\]
The user asks for FCIs:
\begin{align*}
&\textsc{Mining}(\X,\XX,\Y,\YY)=\\
&\begin{cases}
\forall j \in \mathcal{T}: \Y_i=1 \Leftrightarrow \sum\limits_{i\in \mathcal{I}}  \X_i(1-\SDB_{ij})= 0
%\cb{isn't it $\Leftrightarrow$?}
\\
\forall i \in \mathcal{I}: \X_i=1 \Rightarrow \frac{1}{|\mathcal{T}|}\sum\limits_{j\in \mathcal{T}}  \Y_j\SDB_{ij}
%\cb{isn't it Dij?}
\geq \theta\\
\forall i \in \mathcal{I}: \X_i=1 \Leftrightarrow \sum\limits_{j\in \mathcal{T}}\Y_j(1-\SDB_{ij})=0\\
\end{cases}
\end{align*}
This corresponds to the model presented in \cite{guns2011itemset} and how it can be written in the {\sc Mining} part of our {\sc ItemSet} model.
The first constraint represents the coverage constraint, the second is the minimum frequency w.r.t. to a given minimum frequency $\theta$, and the third one expresses the closedness constraint.
Note that to obtain an optimal propagation, this part can be replaced by the global constraint {\sc ClosedPattern}  \cite{Lazaar2016}:
\begin{align*}
\textsc{Mining}(\X,\XX,\Y,\YY)= \textsc{ClosedPattern}_\theta(\X,\Y)
\end{align*}

\subsection*{Query $Q'_1$}
For $Q'_1$, we have $k$ item categories.
The user asks for FCIs extracted from the whole dataset but 
the items composing the extracted FCI must belong 
to at least  $lb$ categories and at most  $ub$ categories 
where $lb\leq ub \leq k$. 
The {\sc DataSet} part is the same as in the case of $Q_1$.
%extracting itemsets from the whole the dataset is written in our model as: 
The {\sc Mining} part takes into account the new user's constraint on itemsets:
\[
\textsc{Mining}(\X,\XX,\Y,\YY)=\begin{cases}
\textsc{ClosedPatten}_\theta(\X,\Y)\\
 lb \leq \sum\limits_{j =1}^k \max\limits_{i \in \mathcal{I}_j}\X_i \leq ub\\
\end{cases}
\]
The first constraint is used to extract FCIs.
The second constraint holds if and only if 
the items of the extracted itemset belong to   
$lb$ to $ub$ categories.

\subsection*{Query $Q_2$}

For $Q_2$, the user asks for FCIs not from the whole dataset as in $Q_1$ and $Q'_1$, 
but from a part of the dataset  with $lb_I$ to $ub_I$  categories of items. 
Such user's constraints on items are expressed in the {\sc DataSet} part of our model as:
\begin{align*}
&  \textsc{DataSet}(\XX,\YY)=\\
&  \begin{cases}
lb_I \leq 
\sum\limits_{j =1}^k \min\limits_{i \in \mathcal{I}_j}\XX_i 
=
\sum\limits_{j =1}^k \max\limits_{i \in \mathcal{I}_j}\XX_i 
\leq ub_I\\
\forall j \in \mathcal{T} :\YY_{j} = 1
\end{cases}
\end{align*}
For each category, the first constraint activates all 
items  or none. The number of categories with their items activated
is between  $lb_I$ to $ub_I$. 
The second constraint  activates the whole set of transactions. 
The {\sc Mining} part is the almost the same as in the case of $Q_1$. The only difference is that we need an adapted version 
of the $\textsc{ClosedPattern}_\theta$ where frequent closed 
itemsets are mined in the sub-dataset circumscribed by 
the $H$ an $V$ vectors:  
 \begin{align*}
 \textsc{Mining}(\X,\XX,\Y,\YY)= \textsc{ClosedPattern}_\theta(\X,\XX,\Y,\YY)
 \end{align*}

\subsection*{Query $Q_3$}
For $Q_3$, we have $v$ transaction categories.
With $Q_3$, the user asks for FCIs not from the whole set of transactions 
but  from at least $lb_T$ and at most $ub_T$ transaction categories. 
 These user's constraints on transactions are written in our model as: 
\begin{align*}
& \textsc{DataSet}(\XX,\YY)=\\
& \begin{cases}
\forall i \in \mathcal{I} : \XX_{i} = 1\\
%  Y'_i = 1 ~\text{s.t}~ i \in \mathcal{T}_j \Rightarrow \forall l \in \mathcal{T}_j,  Y'_l =1 \\
lb_T \leq 
\sum\limits_{j =1}^v \min\limits_{i \in \mathcal{T}_j}\YY_i 
=
\sum\limits_{j =1}^v \max\limits_{i \in \mathcal{T}_j}\YY_i 
\leq ub_T\\
\end{cases} 
\end{align*} 
The first constraint activates the whole set of items. 
For each category, the second constraint activates all  transactions or none. 
The number of categories with their transactions  activated
is between  $lb_T$ and $ub_T$. 
The user asks for CFIs. 
That is, the {\sc Mining} part is the same as in the case of $Q_2$.

\subsection*{Query $Q_4$}
$Q_4$ involves the different types of user's constraints presented in this paper.
	We have $k$ item categories and $v$ transaction categories.
	The user asks for FCIs on at least $lb_I$ and at most $ub_I$ 
	categories of products and at least $lb_T$ and at most $ub_T$ categories of customers. 
\begin{align*}
& \textsc{DataSet}(\XX,\YY)=\\
& \begin{cases}
lb_I \leq 
\sum\limits_{j =1}^k \min\limits_{i \in \mathcal{I}_j}\XX_i 
=
\sum\limits_{j =1}^k \max\limits_{i \in \mathcal{I}_j}\XX_i 
\leq ub_I\\
lb_T \leq 
\sum\limits_{j =1}^v \min\limits_{i \in \mathcal{T}_j}\YY_i 
=
\sum\limits_{j =1}^v \max\limits_{i \in \mathcal{T}_j}\YY_i 
\leq ub_T\\
\end{cases} 
\end{align*} 
The first constraint ensures  the  
sub-dataset satisfies the constraints on items (categories activated as a whole and between $lb_I$ and $ub_I$ of them activated). The second constraint ensures the sub-dataset satisfies the constraints on transactions (categories activated as a whole and between $lb_T$ and $ub_T$ of them activated).
As we look for FCIs, the {\sc Mining} part remains the same as in the case of  $Q_2$ and $Q_3$.

\subsection*{Query $RQ$}
In this section, we illustrate our model on the 
query  presented in Section \ref{sec:ferrari}: {\em Where Ferrari cars are frequently bought?}.
To make it simple, suppose that transactions are categorized 
into $r$ regions 
$\mathcal{T}=\{\mathcal{T}_{1}, \mathcal{T}_{2},\ldots, \mathcal{T}_{r}\}$, 
each region is composed of $d$ departments 
$\mathcal{T}_{i}=\{\mathcal{T}_{i:1}, \mathcal{T}_{i:2},\ldots, \mathcal{T}_{i:d}\}$,  and each department is composed of $c$ cities
$\mathcal{T}_{i:j}=\{\mathcal{T}_{i:j:1}, \mathcal{T}_{i:j:2},\ldots, \mathcal{T}_{i:j:c}\}$. 
(In the real case, the number of cities per department and departments per region can vary.)

The  {\sc Channeling} part of the model is the the same 
as in our generic CP model presented in Section \ref{sec:model}. 

We need now to define the {\sc Dataset} and the {\sc Mining} parts 
for the $RQ$ query. In the following, $f$ refers to 
the item representing the fact that the brand of the car is Ferrari. 
\[
\textsc{DataSet}(\XX,\YY)=\begin{cases}
\forall i \in \mathcal{I}\setminus \{f\} : \XX_{i} = 0\\
H_f=1\\
(1) \vee (2)\vee (3)
\end{cases}
\]
where (1), (2), and (3) are the constraints specifying that itemsets are extracted 
from a region, a department, or a city. 
That is, (1), (2), and (3) are constraints that we can express as in the second line of 
$\textsc{DataSet}(\XX,\YY)$ of query $Q_3$ with 
$lb_T = ub_T = 1$. 

%\begin{align*}
%&(1)\equiv 
%\sum\limits_{i=1}^{r}\max\limits_{o\in\mathcal{T}_i} V_o = 1 \land
%\forall i\in 1..r: \min\limits_{o\in\mathcal{T}_i} \YY_o = \max\limits_{o\in\mathcal{T}_i} \YY_o\\
%&	(2)\equiv
%\sum\limits_{i=1}^{r}\sum\limits_{j=1}^{d}\max\limits_{o\in\mathcal{T}_{i:j}} V_o = 1 \land
% \forall (i,j)\in (1..r,1..d): \min\limits_{o\in\mathcal{T}_{i:j}} \YY_o =  \max\limits_{o\in\mathcal{T}_{i:j}} \YY_o\\
%&	(3)\equiv 
%\sum\limits_{i=1}^{r}
%\sum\limits_{j=1}^{d}
%\sum\limits_{k=1}^{c}\max\limits_{o\in\mathcal{T}_i:j:k} V_o = 1\land 
%\forall (i,j,k)\in (1..r,1..d, 1..c): \min\limits_{o\in\mathcal{T}_{i:j:k}} \YY_o =  \max\limits_{o\in\mathcal{T}_{i:j:k}} \YY_o
%\end{align*}
%
\vspace{-0.4cm}
 \begin{align*}
 \textsc{Mining}(\X,\XX,\Y,\YY)= frequent(\X,\XX,\Y,\YY,10\%)
 \end{align*}
where frequent itemsets are mined in the sub-dataset circumscribed by 
the $H$ an $V$ vectors.

\subsection*{An observation on closedness}

As pointed out in the introduction, closedness  can interfere with  user's constraints 
when they are not  monotone    \cite{BonchiL04}. 
In Example \ref{ex:table}, if we want itemsets  of size 2 
not containing $C,D,E,F$, or $K$, with minimum frequency $30\%$, and 
closed wrt the these constraints, 
no system is able to return the only solution $BG$ because   $BGK$ is closed. 
In our model, $BG$ is returned as a closed itemset of the 
sub-dataset $\mathcal{D}_1[ABCDEFGH]$, that is, when  $H_K=0$.

\section{Experimental Evaluation}\label{sec:expes}

We made experiments to evaluate the queries $Q_1$, $Q_2$, $Q_3$ and $Q_4$ on our generic CP model \itemset{} for itemset mining.

\subsection{Benchmark datasets}

\begin{table}[t] \centering
	\caption{\footnotesize Properties of the used datasets}\label{tab:CharData}
	\scalebox{0.75}{
		\begin{tabular}{|c|r|r|r|r|c|}
			\hline
			Dataset & $|\mathcal{T}|$ & $|\mathcal{I}|$ & $\overline{|\mathcal{T}|}$ & $\rho$ & domain\\
			\hline
			\zoo & 101 & 36  & 16 & 44\% & zoo database \\
			%\hline
			\primary & 336 & 31  & 15 &  48\% & tumor descriptions  \\
			%\hline
			\vote & 435 & 48  & 16 & 33\% & U.S voting Records  \\	
			%\hline
			\chess & 3196 & 75  & 37 & 49\% & game steps  \\
			%\hline
			\mushroom & 8124 & 119  &  23 & 19\% & specie's mushrooms \\
			\hline
			\multicolumn{6}{r}{\primary: Primary-tumor}\\
		\end{tabular}
	}
\end{table}

We selected several real-sized datasets from the FIMI 
repository ({http://fimi.ua.ac.be/data/}) and the CP4IM repository ({https://dtai.cs.kuleuven.be/CP4IM/datasets/}). 
These datasets have various characteristics representing 
different application domains.
For each dataset, Table~\ref{tab:CharData} reports 
the number of transactions $|\mathcal{T}|$, 
the number of items $|\I|$, the average size of transactions 
$\overline{|\mathcal{T}|}$, its 
density $\rho$ (i.e., $\overline{|\mathcal{T}|}/|\I|$), and 
its application domain.
The datasets are presented by increasing  size.
% $|\I|\cdot|\mathcal{T}|$.
%We have selected datasets of various size. 
%The density ranges from $19\%$ to $49\%$.
%The sizes of these datasets vary from $\approx4\times 10^3$ to more than $9.6\times 10^5$.  

\subsection{Experimental protocol}
We  implemented the \itemset{} model presented in Section~\ref{sec:model}. 
This implementation, named \cp, is in C++, on top of the {\tt Gecode} 
solver ({www.gecode.org/}). 
The frequency and closedness constraints are performed by 
a new implementation of the \closedx global constraint 
taking into account the variables $\XX$, $\YY$.
For  \lcm, the state-of-the-art specialized algorithm for CFIs, we 
used the   publicly available version  ({http://research.nii.ac.jp/~uno/codes.htm}). 
All experiments were conducted on an Intel Xeon E5-2665 @2.40 Ghz and a 48\textsc{Gb} RAM with a timeout of $900$ seconds. 

In all our experiments we selected  a minimum support $\theta$ and a 
minimum size of itemsets $k$ in order to have constrained 
instances with less than $10$ solutions. (If a user adds constraints, this to have only 
few but  interesting solutions.)
An instance  is defined by the pair frequency/minsize $(\theta, k)$. 
For example, \zooI{50}{5} denotes the instance of the \zoo{} dataset 
with a minimum support of $50\%$ and solutions of at least $5$ items.
Note that the constraint on the size of the itemset is simply  added to 
the {\sc Mining} part of our \itemset{} model as follows: 
$minSize_{k}(\X)\equiv \sum_{i\in \items}\X_i\geq k$.
Note also that such a constraint is integrated in \lcm{} without the need to 
a post-processing to filter out the undesirable itemsets.

\subsection{Query $Q_1$}

Our first experiment compares \lcm{} and \cp{} on queries of  type $Q_1$,
where we have only user's constraints on itemsets.
We take the $Q_1$ of the example in Section \ref{sec:q1}, 
where the user asks for FCIs.
% (frequency and closedness constraints).
We added the $minSize$ constraint in the {\sc Mining} part of the \itemset model.
Table \ref{tab:resultsQ1} reports the CPU time, in seconds, for each approach on each instance. 
We also report the total number of FCIs  (\FCI$\leq 10$) for each instance.

The main observation that we can draw from Table \ref{tab:resultsQ1} is that, 
as expected, the specialized algorithm \lcm{} wins 
on  all the instances.
However,  \cp{} is quite competitive. % on such constrained instances. 
\lcm{} is only from $1$ to $9$ times faster. 
% than our declarative approach.
%Thanks to CP propagation on such constrained instances. 

\begin{table}[t] \centering
	\caption{\footnotesize \lcm{} and \cp{} on $Q_1$ queries. }\label{tab:resultsQ1}
	\scalebox{0.75}{
		\begin{tabular}{|c|r|r|r|}
			\hline
			\textbf{Instances}	& \FCI{}	&\lcm (sec.)& \cp (sec.) \\
			\hline
			\zooI{50}{5} 		& 4 &  0.01 & 0.01  \\
			%	\hline
			\primaryI{60}{6}	 &1 & 0.01  &  0.01 \\
			%	\hline
			\voteI{50}{2} 		 &1	&  0.01 & 0.01 \\
			%	\hline
			\mushroomI{50}{5}	 & 8 & 0.02 & 0.10  \\
			%	\hline
			\chessI{80}{10} 	 & 4 & 0.03 & 0.29 \\
			
		%	\connectI{90}{12}	& 1 &	0.26 & 2.26 \\
			\hline
		\end{tabular}
	}
\end{table}

% connect 90% minsize=12 -> 1 patterns- > 2.26s 
% Lcm time with size constraint : zoo 0.004, primary : 0.006, vote : 0.005, mushroom : 0.025, chess : 0.022, connect : 0.26

\subsection{Query $Q_2$}

In addition to user's constraints on itemsets, in $Q_2$ the user is able to express 
constraints on items.
We take the $Q_2$ of the example in Section \ref{sec:q2}, 
where  items are in categories and the user asks for FCIs extracted 
from at least $lb_I$ and at most $ub_I$ categories. 
We again added the $minSize$ constraint.

Table \ref{tab:resultsQ2} reports the results of the comparison between \pplcm{} (\lcm{} with a preprocessing) and \cp{} on a set of instances.
For each instance, we report the number of item categories $\#\items_i$, 
the  used $lb_I$ and $ub_I$, the total number $\#D$ of sub-datasets satisfying 
the constraints on items, the number of solutions \FCI{}, and the time in seconds.
Note that the categories have the same size and for a given $\#\items_i=n'$, 
and an $(lb_I, ub_I)$, we have $\#D= \sum_{i=lbI}^{ubI} \binom{n'}{i}$. 

It is important to bear in mind for such a query,  \pplcm
acts in two steps: (i) pre-processing generating all 
possible sud-datasets w.r.t. the user's constraints on items; 
(ii) run \lcm{} on each sub-dataset.
The first step can be very expensive in terms of memory consumption 
because the space complexity of 
generating all sub-datasets is in 
$O(n'\times n\times m)$, where  $n'$ is the number of item categories, 
and $n$ and $m$ the number of items and  transaction.

In  Table \ref{tab:resultsQ2} we observe that \cp{} outperforms  
\pplcm{} on $6$ instances out of $10$. 

% to give an idea of a comparison of \cp\ with 
%the second step of the \lcm{} approach, 
%we report the maximum time that  \lcm{} should spend to be more efficient than \cp{} 
%in terms of CPU time (i.e., $time/\#D$).
%We see that \lcm should be extremely  fast to be competitive, regardless of 
%the space complexity of its preprocessing step. 
%, \cp{} is very competitive with a possible use of \lcm{} after a preprocessing step.

\begin{table}[t] \centering
	\caption{\footnotesize \pplcm{} and \cp{} on $Q_2$ queries.}\label{tab:resultsQ2}
	\scalebox{0.75}{
		
		\begin{tabular}{|c|r|c|r|r|r|r|} \hline
		
			\textbf{Instance} & \#$\items_i$ &  ($lb_I$,$ub_I$)  & $\#D$ & 	\FCI{} & \textbf{(a)} & \textbf{(b)}    \\
			\hline
			\multirow{2}{*}{\zooI{80}{2}} & 6  &  (2,3)  & 35 & 5 & 0.58 & \textbf{0.02}     \\ %6
			& 6  & (3,4)  & 35  & 10 & 0.62 &  \textbf{0.03}    \\%3
			\hline
			\primaryI{70}{5} & 3 &  (2,3) & 4  & 2 & 0.17 & \textbf{0.02}  \\ %5
			\hline
			\voteI{50}{2} & 6 &   (2,3) &  35 & 5 &  0.53 &  \textbf{0.02}   \\ %4
			\hline
			\mushroomI{50}{4} & 17 &   (2,2) &  136 & 9 & \textbf{ 5.32} & 6.14  \\%
			\mushroomI{50}{4}	& 17 &   (2,3)  & 816 & 1  &  \textbf{41.31} &  51.04  \\   %8
			\hline
			\chessI{70}{10} & 5 &  (2,3) &  20  & 1 &  \textbf{1.24} & 2.12  \\%4
			\chessI{80}{10} & 5 &  (2,5) &  26  & 5 & \textbf{1.93} & 3.32   \\%8
			\chessI{70}{5}	& 15 &  (2,2)  & 105 &  6 &  2.78 &  \textbf{0.97 }  \\ %5
			\chessI{80}{6}	& 15 &  (2,3)  & 560 &  2 &  14.57  &  \textbf{7.15}  \\ %1
			\hline
			\multicolumn{7}{r}{\bf (a): \pplcm(s)$\qquad$ (b): \cp(s)}
		\end{tabular}
	}
\end{table}

\subsection{Query $Q_3$}
Let us now present our experiments on queries of type $Q_3$ where we 
have user's constraints on itemsets and transactions.
We take the $Q_3$ of the example in Section \ref{sec:q3} 
where  transactions are in categories.
We added  the $minSize$ constraint.

Table \ref{tab:resultsQ3} reports the results of the comparison between \pplcm{} and \cp{}.
For each instance, we report  the number of transaction categories $\#\trans_i$, 
the lower and upper bounds ($lb_T,ub_T$) on transaction categories, 
the number of sub-datasets $\#D$, the number of extracted solutions \FCI{} and the time in seconds.
Note that for a number  of categories $\#\trans_i=m'$ and a given ($lb_T,ub_T$), we have 
$\#D= \sum_{i=lbT}^{ubT} \binom{m'}{i}$.

For $Q_3$,  \pplcm acts again in two steps. 
The space complexity of the preprocessing step is in   
$O(m'\times n\times m)$, with $m'$ transaction categories, $n$ items and $m$ transactions.

In Table \ref{tab:resultsQ3} we observe  that 
\cp{} is faster than \pplcm{} on 6 instances out of 10. 
\cp wins on instances where $\#D$ is large.  
On \voteI{80}{3} with $\#\mathcal{T}_i=29$ and $(lb_T,ub_T)=(2,5)$,  
\pplcm reports a timeout whereas \cp solves it in 12 minutes. 
%Let us take the instance \voteI{80}{3}, with $29$ transaction categories 
%and $(lb_T,ub_T)=(2,5)$. The $146,566$ sub-datasets need more than one GB.
%For \chessI{90}{26}, we need more than $14$GB.

\begin{table}[t] \centering
	\caption{\footnotesize \pplcm{} and \cp{} on $Q_3$ queries. }\label{tab:resultsQ3}
	\scalebox{0.75}{
		
		\begin{tabular}{|c|r|c|r|r|r|r|}\hline
			\textbf{Instance} & \#$\trans_i$ &  ($lb_T$,$ub_T$)  & $\#D$ & \FCI{} & \textbf{(a)} & \textbf{(b)}    \\
			\hline
			\zooI{70}{10} & 10  &  (1,10)  & 1,023 & 2 &  7.95 &  \textbf{1.12}   \\ %1
			\zooI{80}{5} & 10  & (2,10)  &  1,013 & 8 &  9.05 &  \textbf{1.37}    \\ %9
			\hline
			\primaryI{85}{4} & 7 &   (2,7)  & 120 & 1  &  1.45 & \textbf{0.25}    \\ %8
			\hline
			\voteI{70}{6} & 29 &   (2,3) &  4,060 & 3 &  37.93 & \textbf{17.95}   \\ %3
			\voteI{80}{3}& 29  &  (2,4) & 27,811   & 4  &  324.53&  \textbf{135.53}   \\ %9
			\voteI{80}{3} & 29  &  (2,5) &   146,566 & 4  &  \textsc{to} &  \textbf{739.31}  \\
			\hline
			\multirow{2}{*}{\mushroomI{70}{12}} & 12 &   (2,2) & 66 & 3  &  \textbf{3.13} &  24.45   \\ %6
			& 12 &   (3,3)  & 220 &  2 &  \textbf{12.63} &  87.65  \\         %1
			\hline
			\chessI{90}{22}	  & 34 &   (2,2) &  561 & 1 &  \textbf{8.43} & 15.10  \\ %9
			\chessI{90}{26} & 94   &  (2,2)  & 4,371 & 3  &  \textbf{49.73} &  68.82  \\ %2
			\hline
						\multicolumn{7}{r}{\bf {\sc to}: timeout $\qquad$ (a): \pplcm(s)$\qquad$ (b): \cp(s)}
			
		\end{tabular}
	}
\end{table}

\subsection{Query $Q_4$}

Our last experiment is on queries of type $Q_4$ where the user can put constraints 
on both items and transactions in addition to the ones on the itemsets themselves.
We take the $Q_4$ of the example in Section \ref{sec:q4} 
where  items and transactions are  in categories. 
We added  the $minSize$ constraint.

Table \ref{tab:resultsQ4} reports results of the comparison between \pplcm{} and \cp{} acting on different instances. 
We report the number of uniform categories of items/transactions, the used $(lb_I,ub_I)$ and $(lb_T,ub_T)$, the number of sub-datasets  $\#D$, the number of solutions \FCI{} and the time in seconds.
\pplcm{} needs to generate all possible sub-datasets 
$\#D= \sum_{i=lbT}^{ubT} \binom{m'}{i} \times \sum_{i=lbI}^{ubI} \binom{n'}{i}$, 
where  $n', m', n$ and $m$ are respectively the number of item categories,  transaction categories,  items and  transactions.
\cp{} is able to deal with the different queries $Q_4$ 
just by changing the parameters $k, lb_T, ub_T, lb_I, ub_I$, whereas \pplcm{} needs a time/memory consuming preprocessing before each query.

We see in Table \ref{tab:resultsQ4} that \cp{} significantly outperforms \pplcm{}.
On the instances where  \pplcm{} does not report a timeout, 
\cp{} is from $4$ to more than $26$ times faster than \pplcm{}.
The pre-processing step of \pplcm{} can reach $90\%$ of the total time.
As  $\#D$  grows exponentially, it quickly 
leads to an infeasible preprocessing step (see the 3 timeout cases of \pplcm).

%
%Let us take \zoo\ instances. 
%We observe that the time required by \cp{} seems to increase 
%sub-exponentially in the number  $\#D$ of sub-datasets. 

\begin{table}[t] \centering
	\caption{\footnotesize \lcm{} and \cp{} on $Q_4$ queries. }\label{tab:resultsQ4}
	\scalebox{0.61}{
		
		\begin{tabular}{|c|r|r|c|c|r|r|r|r|}
			\hline
			\textbf{Instances} & $\#\items_i$ & $\#\trans_i$ & ($lb_I$,$ub_I$)  & ($lb_T$,$ub_T$) &  $\#D$ & \FCI{} &\textbf{(a)} & \textbf{(b)}   \\
			\hline
			\zooI{70}{6} & 6 & 10  &  (2,3) & (2,3) & 5,775 & 8  & 39.69 &\textbf{1.75}   \\ %1
			\zooI{50}{11} & 6 & 10  & (3,4) & (3,4)  & 11,550 &  9 &  88.66 & \textbf{3.36}  \\ %6
			\zooI{85}{5} & 6 & 10  & (2,6) & (2,10) & 57,741 & 8  &  521.89 & \textbf{31.86} \\ %2
			\hline
			\primaryI{82}{5}& 3 & 12  & (2,3) & (2,10) & 16,280 & 8  &  199.58 & \textbf{36.13}  \\
			\hline
			\voteI{70}{6} & 6 & 29  &  (2,3) & (2,3) & 142,100 & 2 &  \textsc{to} & \textbf{118.67}  \\
			\voteI{72}{5} & 8 & 29  &  (2,3) & (2,3) & 341,040 & 2  &  \textsc{to} & \textbf{201.79}  \\
			\hline
			\mushroomI{80}{5} & 17 & 12  &  (2,2) & (2,2) & 8,976 & 10 & 446.42  & \textbf{102.68}  \\ %2
			\mushroomI{82}{5} & 17 & 12  &  (2,2) & (3,3) & 29,920 & 7  &  \textsc{to} & \textbf{455.19}  \\         
			\hline
			\chessI{90}{16}	 & 5 & 34  &  (2,3) & (2,2) & 11,220 & 3 & 286.42  & \textbf{87.22}  \\ %2
			\hline
									\multicolumn{9}{r}{\bf {\sc to}: timeout $\qquad$ (a): \pplcm(s)$\qquad$ (b): \cp(s)}
			
		\end{tabular}
	}
\end{table}

%\mm{Example for memory: in Q4 Primary needs 121,3 Mo and Mushroom1 needs 140,1 Mo, Mushroom2 if we go far than TO (2685.94s) needs 613,4 Mo}
%
%\mm{Mettre à jour le TO dans le texte à 900}

\section{Conclusion}
%In this paper we have presented a general CP model to catch any kind of user's constraints presented as a taxonomy of constraints. 
%The user can express constraints on the dataset in terms of items and transactions 
%before the mining process. We have illustrated each type of query using our CP model.

We have presented a taxonomy of the different types of user's constraints for itemset 
mining. 
Constraints  can  express properties on the itemsets as well as on the items and transactions 
that compose the datasets on which to look. 
We have introduced a generic constraint programming model for itemset mining.
%In our model, queries may involve any type of user's constraints defined in the taxonmy. 
%A query  can thus specify what kind of  itemsets to find in what kind of sub-datasets.
%
We showed  how our generic CP model can easily take into account any type of user's constraints.
We empirically evaluated
our CP model. We have shown that it can handle the different types of constraints on different datasets. 
The CP approach can find the itemsets satisfying all user’s constraints
in an efficient way compared to the specialized algorithm  \lcm{}, which requires
a memory/time consuming preprocessing step.

\bibliographystyle{plain}
%\cb{references to be made homogeneous}

%\bibliography{biblio}

\end{document}